\documentclass[10pt,twocolumn,letterpaper]{article}

\usepackage{iccv}
\usepackage{times}
\usepackage{epsfig}
\usepackage{graphicx}
\usepackage{amsmath}
\usepackage{amssymb}
\usepackage{fancyhdr}
\usepackage{footmisc}

\fancypagestyle{firstpage}{
    \fancyhf{}
    \fancyfoot[C]{\small International Conference on Computer Vision (ICCV) - HANDS Workshop, 2023}
}
\usepackage[breaklinks=true,bookmarks=false]{hyperref}

\iccvfinalcopy 

\def\iccvPaperID{****} 

\ificcvfinal\fi

\begin{document}

\title{New keypoint-based approach for recognising British Sign Language (BSL) from sequences}

\author{Oishi Deb \thanks{Corresponding author: oishideb@robots.ox.ac.uk}\\
Visual Geometry Group (VGG)\\
University of Oxford\\
\and
KR Prajwal\\
Visual Geometry Group (VGG)\\
University of Oxford\\
\and
Andrew Zisserman\\
Visual Geometry Group (VGG)\\
University of Oxford\\
}

\maketitle
\ificcvfinal\thispagestyle{empty}\fi
\pagestyle{fancy}
\thispagestyle{firstpage}

\begin{abstract}
   In this paper, we present a novel keypoint-based classification model designed to recognise British Sign Language (BSL) words within continuous signing sequences. Our model's performance is assessed using the BOBSL dataset, revealing that the keypoint-based approach surpasses its RGB-based counterpart in computational efficiency and memory usage. Furthermore, it offers expedited training times and demands fewer computational resources. To the best of our knowledge, this is the inaugural application of a keypoint-based model for BSL word classification, rendering direct comparisons with existing works unavailable.
\end{abstract}

\section{Introduction}

Sign languages, having evolved within deaf communities, are unique visual languages with rich grammatical structures and lexicons, often displaying complexities similar to spoken languages. In our study, we introduce an innovative keypoint-based approach for recognizing words from British Sign Language (BSL), which is the primary language used within the British deaf community.

Our research aims to leverage the power of keypoint representations, which provide valuable insights into the body gestures and expressions used in BSL. By utilizing this approach, we seek to enhance the accessibility and efficiency of BSL recognition, contributing to improved communication for the deaf community. The dataset used in our study encompasses a large-scale collection of diverse BSL expressions, enabling comprehensive training and evaluation of our proposed model. Through this work, we hope to make significant strides in promoting computationally efficient models for British sign language recognition and fostering inclusivity for individuals with hearing impairments, as leveraging AI technologies can drive enhanced inclusivity across various domains \cite{responseAI_report}, including support for individuals with special needs or disabilities.

Instead of directly utilizing RGB videos, we explore the use of 2D keypoints, which represent specific points in the face, right hand, left hand and pose. There are three primary reasons for adopting this approach. Firstly, we can easily control the information used by including or excluding subsets of keypoints. For example, instead of using a facial mesh, we can easily reduce the number of facial keypoints to only eyes, lips and the outer face circumference, and therefore reducing the computational cost. Secondly, keypoints eliminate many extraneous factors, such as lighting and clothing, providing a more compact representation compared to images. Consequently, training and exploring models become more manageable. Lastly, keypoints can be computed at frame rate, enabling real-time execution of models for the task.

 Recognising signing words from continuous signing clips is a very challenging task because of co-articulated signing, also known as "signs in context," refers to signing that demonstrates variations in sign form influenced by the signs that come immediately before or after, or signs articulated simultaneously. To develop resilient models capable of comprehending sign language in real-world settings, it is essential to effectively recognise and account for co-articulated signs.

In this paper, our primary contribution is the development of a keypoint-based model designed to identify words from given BSL signing clips. Our evaluations indicate that this keypoint-centric approach substantially outperforms the RGB-based model in computational efficiency, memory usage, and training speed, thus demanding fewer overall computational resources.

\section{Literature Review}
\subsection{Sign Recognition}
There have been multiple works \cite{BullEtAl2020} on British Sign Language recognition. \cite{albanie2021bsl1k} presents an approach that aims to recognise coarticulated signs, which are signs that blend or flow into each other, using mouthing cues. Recognising that signs are often influenced or modified by the simultaneous mouthing of English words, the study incorporates these cues to scale up recognition systems. \cite{adeyanju2021mlmethods} explores the importance of lip patterns in understanding sign language. BSL, like many sign languages, incorporates facial expressions and mouth movements as integral components. This study aims to leverage machine learning techniques to accurately recognise and interpret these lip patterns. 

However, all the previous work uses RGB videos \cite{prajwal2022weaklysupervised}, this work takes a different approach using keypoint representations, as explained below.

\subsection{Keypoint Representation}
In recent years, there has been a shift in the methods used for estimating human body keypoints. Initially, manual feature engineering was the primary approach, but more recent advancements have embraced deep learning architectures such as Convolutional Neural Networks (CNNs) to estimate more accurate keypoints for human-body \cite{CaoEtAl2019, GulerEtAl2018, LinnaEtAl2016, LugaresiEtAl2019, PfisterEtAl2015}. Various studies have explored the utilization of keypoints as inputs for different tasks, and one popular approach involves constructing Graph Neural Networks (GNNs) based on keypoint representations.

GNNs have been applied in tasks like action recognition \cite{YanEtAl2018}, gesture recognition \cite{LiEtAl2021}, and sign language segmentation and recognition \cite{BullEtAl2020, AmorimEtAl2019, MengLi2021, VazquezEnriquezEtAl2021}, utilizing keypoints as inputs. These studies have demonstrated remarkable outcomes, highlighting the effectiveness of employing keypoint representations. In contrast, our research takes a slightly different approach by directly utilizing keypoint representations as inputs to a Transformer-based model.

The Transformer architecture \cite{vaswani2023attention}, originally proposed for natural language processing tasks, has gained prominence in computer vision applications. By leveraging self-attention mechanisms, Transformers are capable of capturing long-range dependencies and modeling relationships between keypoints effectively. Our work capitalizes on this potential by employing keypoint representations as direct inputs to the Transformer model.

\section{Procedure}

\subsection{Dataset}
BOBSL dataset, which contains a total of 1,467 hours of BBC episodes, was used in this work.
Out of various other available sign language datasets \cite{Albanie2021bobsl}, we chose the BOBSL dataset due to the following advantages, unlike datasets with isolated signs, BOBSL comprises co-articulated signs, representing a more natural signing style (although distinct from conversational signing due to its use of interpreted content). Additionally, BOBSL is the largest dataset in terms of continuous signing, with a total duration of 1,467 hours. It covers a broad domain of discourse and benefits from automatic annotation for a substantial vocabulary of 8162 signs.

Each episode is 30 to 60 minutes long, and the videos in the dataset have a resolution of 444 × 444 pixels and a frame rate of 25 fps.

The dataset also includes approximately 1.2 million sentences extracted from English subtitles, which cover a vocabulary size of 78,000 English words. It involves a total of 39 signers (interpreters) \cite{Albanie2021bobsl}. To facilitate signer-independent evaluation, we divide the data into three splits: train, validation, and test, ensuring that there is no overlap of signers between these splits.

This is a classification problem with 8162 word categories, we have used 8162 unique words for training which corresponds to a total of 3,555,141 frames, and we used 3348 words for the validation set corresponding to a total of 53,768 frames.

\subsection{Keypoints Extraction}

In order to obtain the keypoints, we employ Mediapipe \cite{LugaresiEtAl2019}, a publicly accessible library designed specifically for human pose estimation. The main benefit of utilizing Mediapipe, as opposed to alternative libraries like OpenPose \cite{CaoEtAl2019}, is its capability to extract keypoints in real time, even when running on a CPU. This feature significantly reduces the computational load associated with the process. We extracted 33 pose keypoints, 21 keypoints each for left and right fingers, and finally extracted 468 face keypoints.

In most cases, only one person (the BSL interpreter) is present in the frames. In the rare instances where additional people are visible in the background, we only extract the keypoints of the person with the highest confidence using the Mediapipe framework. This ensures that we process a maximum of one signer per frame. Some examples of multiple people in the frame are shown in the appendix. We extract keypoints for about 132 Millions frames in total.

\subsection{Implementation Details}

First, we extract keypoints from every frame, captured at a frame rate of 25 frames per second. We feed a sequence of 16 consecutive keypoint vectors corresponding to frames into the model. Prior research \cite{ViitaniemiEtAl2014, BuehlerEtAl2009, PfisterEtAl2013} has observed that co-articulated signs, often referred to as "signs in context," tend to have a duration of approximately 13 - 20 frames.

The keypoints, comprising (x, y) coordinates, from 16 consecutive frames are stacked together to form a $[16 \times K \times 2]$ three-dimensional vector (where K represents the total number of keypoints per frame). This keypoint vector is then fed into the transformer for sign recognition. Further details of the Transformer architecture are in the next section.

\subsubsection{Transformer Architecture}

\begin{figure*}[htbp]
\centerline{\includegraphics[width= 11.5 cm, height=4cm]{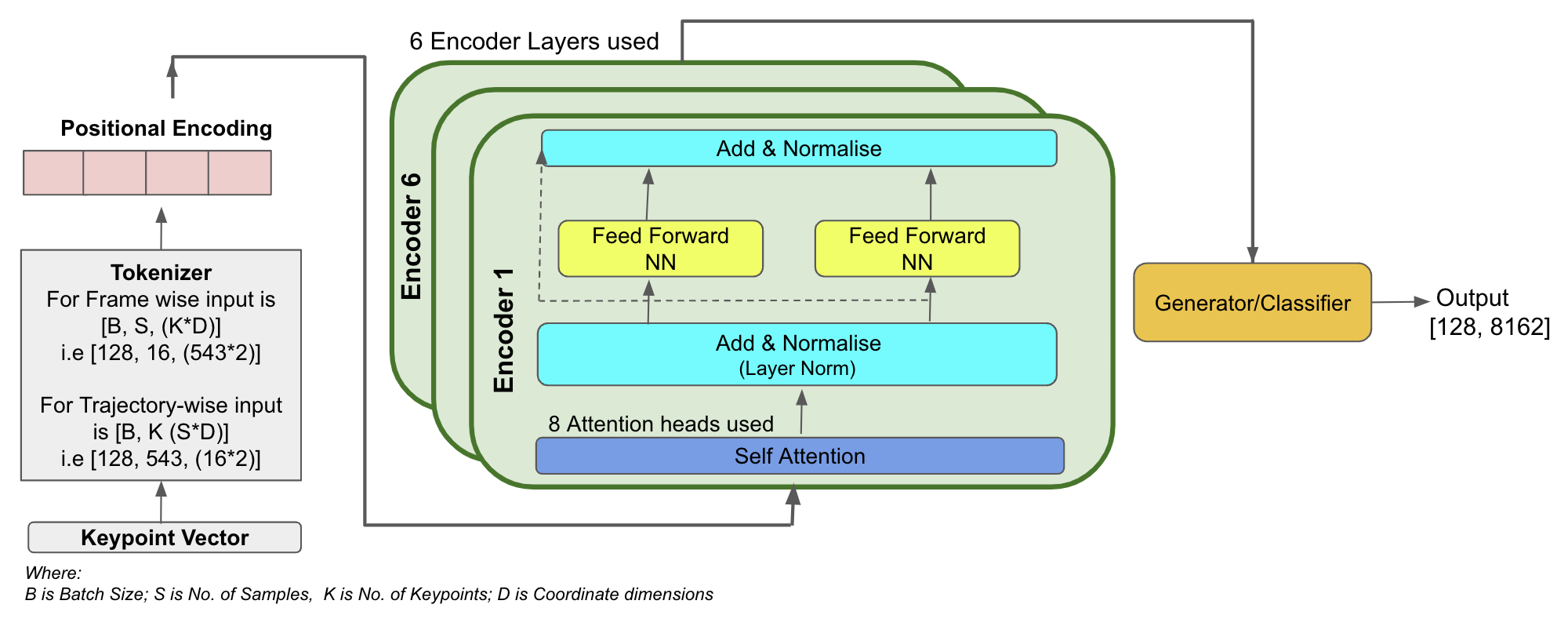}}
\caption{Transformer Architecture for Frame-wise and Trajectory-wise attention model.}
\label{archi.png}
\end{figure*}

The Transformer model is composed of several integral components: the Tokenizer, Positional Encoding, Encoder, Multi-Headed Attention, Position-wise Feed Forward neural network, and a Generator, illustrated by Figure \ref{archi.png}.

Initially, the input data is tokenized by the Tokenizer and supplemented with positional information via the Positional Encoding module. This encoding allows the model to discern the order and position of tokens within the sequence.

Subsequently, the transformed input navigates through the Encoder, which employs self-attention and feed-forward mechanisms to derive context-rich representations. Each Encoder comprises successive layers of self-attention and feed-forward neural network, facilitating the model's grasp of hierarchical and context-dependent aspects of the data. A total of eight attention heads have been used in each encoder module, which employs the scaled dot-product attention mechanism, discerning dependencies across varying sequence positions. Meanwhile, the Position-wise Feed Forward network applies a feed-forward neural network individually to each sequence position. Layer normalization is executed via the Layer-Norm functionality to ensure stable training.

The culmination of this process sees the Generator module transforming the learned representations into the specified class outputs.

\subsubsection{Hyperparameters}
Our Transformer model consists of six encoder layers, each equipped with eight attention heads. Our model uses 512-dimensional embeddings. To train our network, we utilize the Adam optimizer \cite{KingmaBa2014} with a learning rate set to 1e-4. We use a batch size of 128, and the training process is halted when the validation loss does not improve for 3 consecutive epochs.

\section{Evaluation and Discussions}

\subsection{Experiments}
We conducted training using various methodologies, including the Frame-wise Attention model and the Trajectory-wise Attention model, without employing any data augmentation. During our experiments, we explored models with different counts of total keypoints, specifically 543 and 203 keypoints (kp). In the 543 kp model, we used 468 keypoints in the face, whereas in the 203 kp model we used 128 face keypoints. Interestingly, we observed enhanced performance when increasing the number of keypoints associated with facial features.

Building on these findings, our experimentation extended to data augmentation techniques, as elaborated in the subsequent section. 

\subsubsection{Augmentation}
We explored several augmentation techniques including shifting, scaling, rotating, and horizontal flipping. Figure \ref{shift_aug.png} below illustrates the shift augmentation, while the illustrations for the remaining techniques are detailed in the Appendix. Model performance improved with each of the augmentation methods when trained with each separately, and training with shift augmentation provided the maximum boost to the performance.

\begin{figure*}[htbp]
\centerline{\includegraphics[width= 12.5 cm, height=4.5cm]{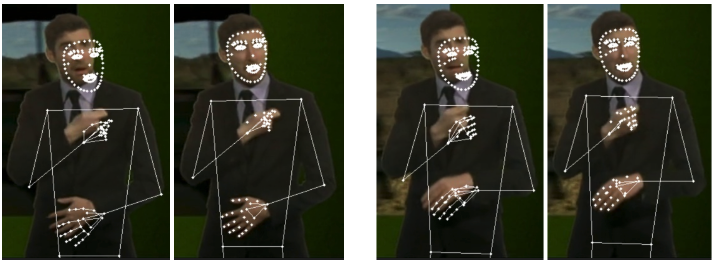}}
\caption{Shift augmentation is exemplified on two distinct frames, including the original keypoint representation prior to augmentation. For visualization clarity, the shift is depicted as 15 coordinates on both the x and y-axis. However, within the code, shifts range between -2 and 2. It's pertinent to note that while RGB images are overlaid for visualization, they are not utilized in model training.}
\label{shift_aug.png}
\end{figure*}

\subsection{Evaluation}
\subsubsection{Accuracy}
The top-5 percent accuracy has been calculated for all our models, which is 60\%. Considering the inherent challenges of this task, this performance is commendable, particularly since we solely rely on keypoint vectors without incorporating RGB images. To the best of our knowledge, ours is the first work in only using keypoint vectors for this task, hence comparison with other models is infeasible.
On the upside, we've notably reduced computational costs, training duration, and memory usage. There's potential for even greater accuracy by integrating keypoints with RGB images.

\subsubsection{Computational Comparison}
Our model employs 23.9 million parameters, compared to the 34.5 million parameters associated with RGB images. By exclusively utilizing keypoint vectors, we achieve a significant reduction in computational cost.

\section{Conclusion and Further Work}
Although our model demonstrates 60\% accuracy on unseen data, but its computational efficiency is noteworthy. This is the pioneering effort to utilize a keypoint-based model for British Sign Language word classification, rendering direct comparisons with existing works unavailable. We acknowledge the potential for refining our model to enhance accuracy. Future extensions might incorporate keypoint-based imagery, specifically black-and-white skeleton images. Additionally, we plan to explore advanced sequence-level 3D pose estimation techniques in subsequent research.

\section*{Acknowledgement}
 This research was funded by the UKRI's EPSRC AIMS grant (EP/S024050/1), and we appreciate their support.

{\small
\bibliographystyle{ieee_fullname}
\bibliography{reference}
}

\newpage
\onecolumn
\section*{APPENDIX}
\appendix
\begin{figure*}[htbp]
\centerline{\includegraphics[width= 11.5 cm, height=3.5cm]{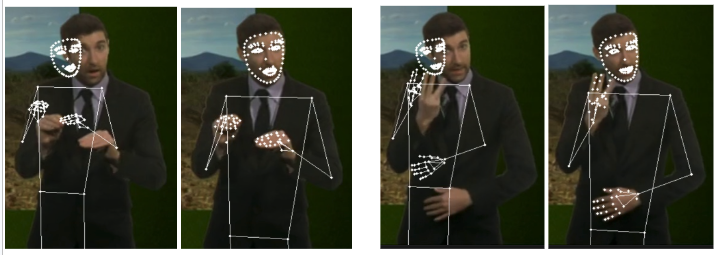}}
\caption{Scale augmentation is illustrated using two distinct frames, with the original keypoint representation also displayed before augmentation. For visualization, the scale is depicted as 75 percent. However, in the actual code, scaling varies between 90 to 110 percent.}
\label{scale_aug.png}
\end{figure*}

\begin{figure*}[htbp]
\centerline{\includegraphics[width= 11.5 cm, height=3.5cm]{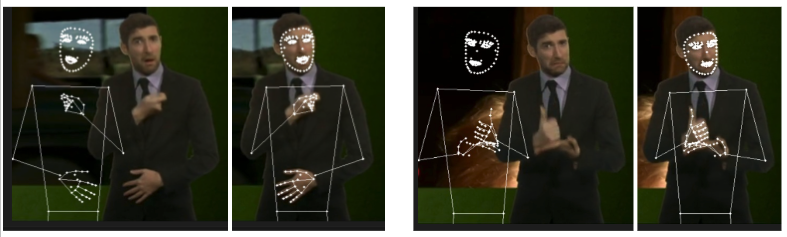}}
\caption{Horizontally Flipped augmentation example on two different frames, it also shows the original keypoint representation before the augmentation.}
\label{hori_aug.png}
\end{figure*}

\begin{figure*}[htbp]
\centerline{\includegraphics[width= 11.5 cm, height=3.5cm]{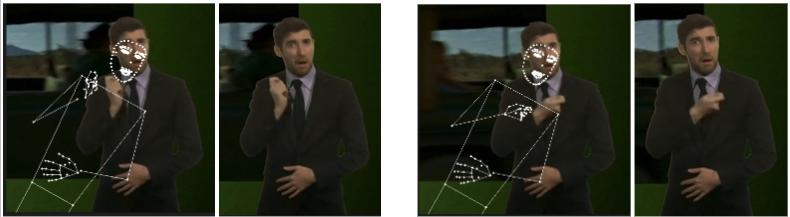}}
\caption{Rotated augmentation example on two different frames, it also shows the original keypoint representation before the augmentation.}
\label{rotated_aug.png}
\end{figure*}

\begin{figure*}[htbp]
\centerline{\includegraphics[width= 11.5 cm, height=4.5cm]{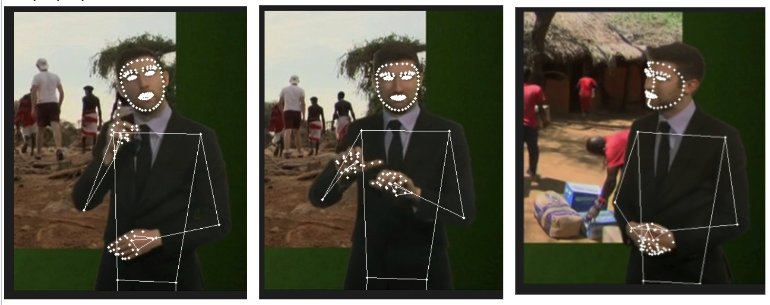}}
\caption{Here are some examples of frames with multiple people, we only extract the keypoints from the signer.}
\label{multiple_people_frames.png}
\end{figure*}

\end{document}